\documentclass[runningheads]{llncs}
\usepackage[T1]{fontenc}
\usepackage{amsmath}
\usepackage{graphicx}
\usepackage{newtxtext,newtxmath}
\usepackage{cite}
\usepackage{amsmath}
\usepackage{algorithmic}
\usepackage{textcomp}
\usepackage{xcolor}
\usepackage{booktabs}
\usepackage{float}
\usepackage{multirow}
\begin{document}
\title{ALADIN:Attribute-Language Distillation Network for Person Re-Identification}

\titlerunning{ALADIN:Attribute-Language Distillation Network for Person Re-Identification}

\author{Wang Zhou\inst{1} \textsuperscript{$\star$} \and
Boran Duan\inst{1}\thanks{Equal contribution.} \and
Haojun Ai\inst{1} \thanks{Corresponding author.} \and Ruiqi Lan\inst{1} \and Ziyue Zhou\inst{1}}
\authorrunning{W.Zhou et al.}
%
\institute{Wuhan University No.299 Bayi Road, 430072, PR China}

\maketitle              
\begin{abstract}
Recent vision–language models such as CLIP provide strong cross-modal alignment, but current CLIP-guided ReID pipelines rely on global features and fixed prompts. This limits their ability to capture fine-grained attribute cues and adapt to diverse appearances. We propose ALADIN, an attribute–language distillation network that distills knowledge from a frozen CLIP teacher to a lightweight ReID student. ALADIN introduces fine-grained attribute–local alignment to establish adaptive text–visual correspondence and robust representation learning. A Scene-Aware Prompt Generator produces image-specific soft prompts to facilitate adaptive alignment. Attribute-local distillation enforces consistency between textual attributes and local visual features, significantly enhancing robustness under occlusions. Furthermore, we employ cross-modal contrastive and relation distillation to preserve the inherent structural relationships among attributes. To provide precise supervision, we leverage Multimodal LLMs to generate structured attribute descriptions, which are then converted into localized attention maps via CLIP. At inference, only the student is used. Experiments on Market-1501, DukeMTMC-reID, and MSMT17 show improvements over CNN-, Transformer-, and CLIP-based methods, with better generalization and interpretability. 
\end{abstract}
\section{Introduction}
\label{sec:intro}
Person Re-Identification (ReID) aims to match pedestrian images captured by different cameras.
It is a key task in intelligent video surveillance.Traditional visual-only ReID methods\cite{Luo_2019_CVPR_Workshops,sun2018PCB,He_2021_ICCV} degrade under occlusion, appearance changes, and domain shifts. Recent vision–language models such as CLIP \cite{Radford2021CLIP} offer strong image–text alignment. They also encode attribute-level semantics, which help distinguish visually similar identities.

CLIP-ReID \cite{DBLP:conf/aaai/LiSL23} first applied CLIP to person ReID. It showed that vision–language pre-training improves identity matching across domains. Following this, several studies further enhance CLIP-based representation learning: PromptSG \cite{Yang2024PromptSG} adapts CLIP through learnable soft prompts, demonstrating the benefit of textual guidance; CLIP3DReID \cite{Liu2024CLIP3DReID} distills body-shape semantics to improve part-level robustness; and IDEA \cite{wang2025idea} fuses textual and visual cues via deformable aggregation. 
However, most existing methods adopt fixed, handcrafted prompts that ignore image-specific appearance changes. As a result, they fail to capture fine-grained attributes that are essential for distinguishing visually similar pedestrians.

\begin{figure}[!t]
    \centering
    \includegraphics[width=\linewidth]{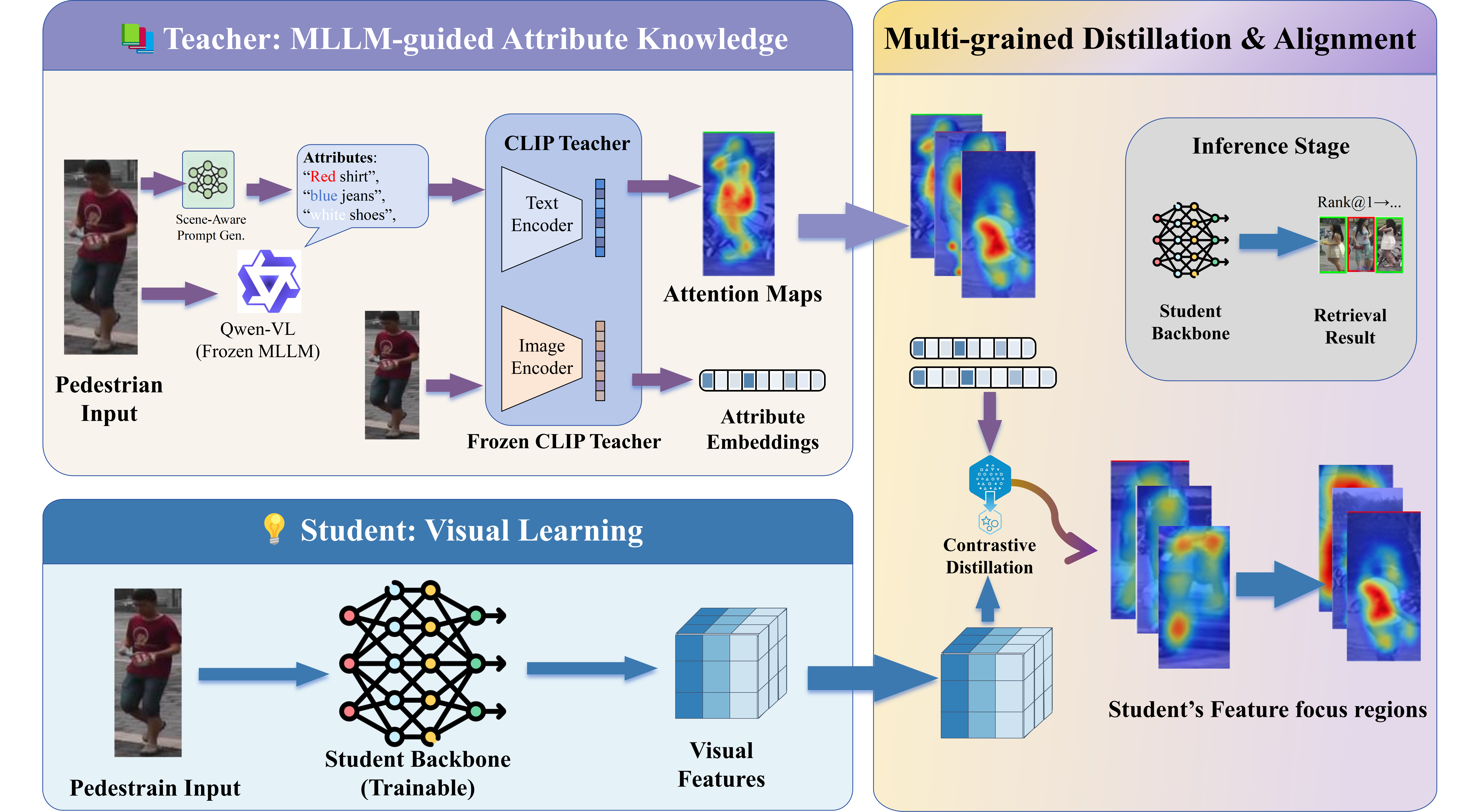}
    \caption{\fontsize{8pt}{9pt}\selectfont
    Overview of the proposed attribute–language distillation framework. Given an input image, an MLLM generates structured attribute descriptions, which are parsed into semantic groups and encoded by the frozen CLIP text encoder with scene-aware soft prompts. The CLIP image encoder provides attention maps and attribute vectors that guide multi-level distillation. These semantic cues are transferred to a lightweight student ReID backbone, enabling fine-grained attribute alignment and robust identity representation.}
    \label{fig:framework}
\end{figure}

To tackle these limitations, we propose ALADIN, an attribute–language distillation network for person ReID.
It uses a multimodal LLM (MLLM, e.g., Qwen-VL \cite{Qwen-VL}) to generate appearance descriptions without manual annotation. As shown in Fig.~\ref{fig:framework}, the MLLM produces structured attribute sentences describing colors, clothing, and accessories. A frozen CLIP teacher encodes these descriptions to produce attention maps and attribute features, guiding the model to focus on the correct visual regions. The student network learns from these signals by aligning each attribute with the corresponding image region and distilling knowledge through contrastive and relational learning. A scene-aware prompt module further adjusts text prompts according to the input image.

Our method combines text descriptions from the MLLM with attention from CLIP. This gives the student automatic and more interpretable supervision. This integrated design enables the student to acquire fine-grained attribute knowledge without relying on handcrafted prompts. Only the student network is used for inference, reducing memory usage and speeding up deployment.

Our contributions are summarized as follows:
\begin{itemize}
\item We introduce an attribute–language distillation framework for person ReID, which transfers fine-grained attribute cues from a frozen CLIP teacher to a lightweight visual model, helping the student learn discriminative details of  appearance.

\item We design a scene-aware soft prompt module that dynamically generates image-specific prompts conditioned on visual features.
By adapting the prompts to scene and appearance variations across camera views, this module enables more stable image–text alignment and consistently improves retrieval accuracy.

\item During inference, only the student network is required, without any text or multimodal modules.
This keeps the computation and memory cost similar to a basic visual ReID model, making deployment easier.

\end{itemize}

\section{Related Works}

\subsection{CLIP-based Person Re-Identification}
Vision–language models such as CLIP \cite{Radford2021CLIP} have inspired many recent ReID methods by providing strong cross-modal supervision.
CLIP-ReID \cite{DBLP:conf/aaai/LiSL23} first demonstrated that CLIP’s semantic priors can benefit identity discrimination even without explicit text labels.
Subsequent works explore different ways of leveraging CLIP: PromptSG \cite{Yang2024PromptSG} employs learnable prompts to enhance attention, while CLIP3DReID \cite{Liu2024CLIP3DReID} distills 3D body-shape cues via linguistic tokens.
In addition, several studies investigate CLIP-driven domain generalization or cross-modal transfer for ReID, such as visible–infrared alignment \cite{zhang2024clipvisinfra} and multi-modal robust features \cite{wang2025idea}. 
Despite these advances, most CLIP-based methods use only global features or fixed prompts.
As a result, they often miss important local details and struggle to tell similar pedestrians apart.

\subsection{Prompt-, Attribute-, and Text-Enhanced ReID}

Prompt learning (e.g., CoOp \cite{zhou2022coop}) is widely used to adapt VLMs.
In ReID, prompt-based methods incorporate structural priors such as human keypoints \cite{DBLP:conf/eccv/SomersAV24} or identity prototypes \cite{li2023prototypical}.
Attribute-aware approaches like AMD \cite{chen2021AMD} distill appearance cues, and modality-confusion or aggregation designs \cite{hao2021cmreid} reduce cross-modality gaps.
Dual-attention and refinement strategies \cite{DBLP:conf/cvpr/ZhuKLLTS22} further indicate the importance of fine-grained region modeling.

Text supervision also brings complementary semantics.
TIP-CLIP \cite{yan2023tipclip} aligns images with detailed descriptions, UniPT \cite{shao2023unipt} uses pseudo texts for unified pre-training, and MLLM-based methods \cite{tan2024harnessing} generate attribute sentences for training.
Multimodal techniques such as Magic Tokens \cite{Zhang_2024_CVPR} and sequence-level fusion \cite{yu2024tfclip} enhance robustness through token selection or temporal modeling.

However, most existing methods rely on shallow prompt tuning, handcrafted or noisy attributes, and predominantly global alignment. These limitations reduce their ability to capture diverse fine-grained cues and weaken the reliability of cross-modal correspondence.

\begin{figure}[t]
    \centering
    \includegraphics[width=\linewidth]{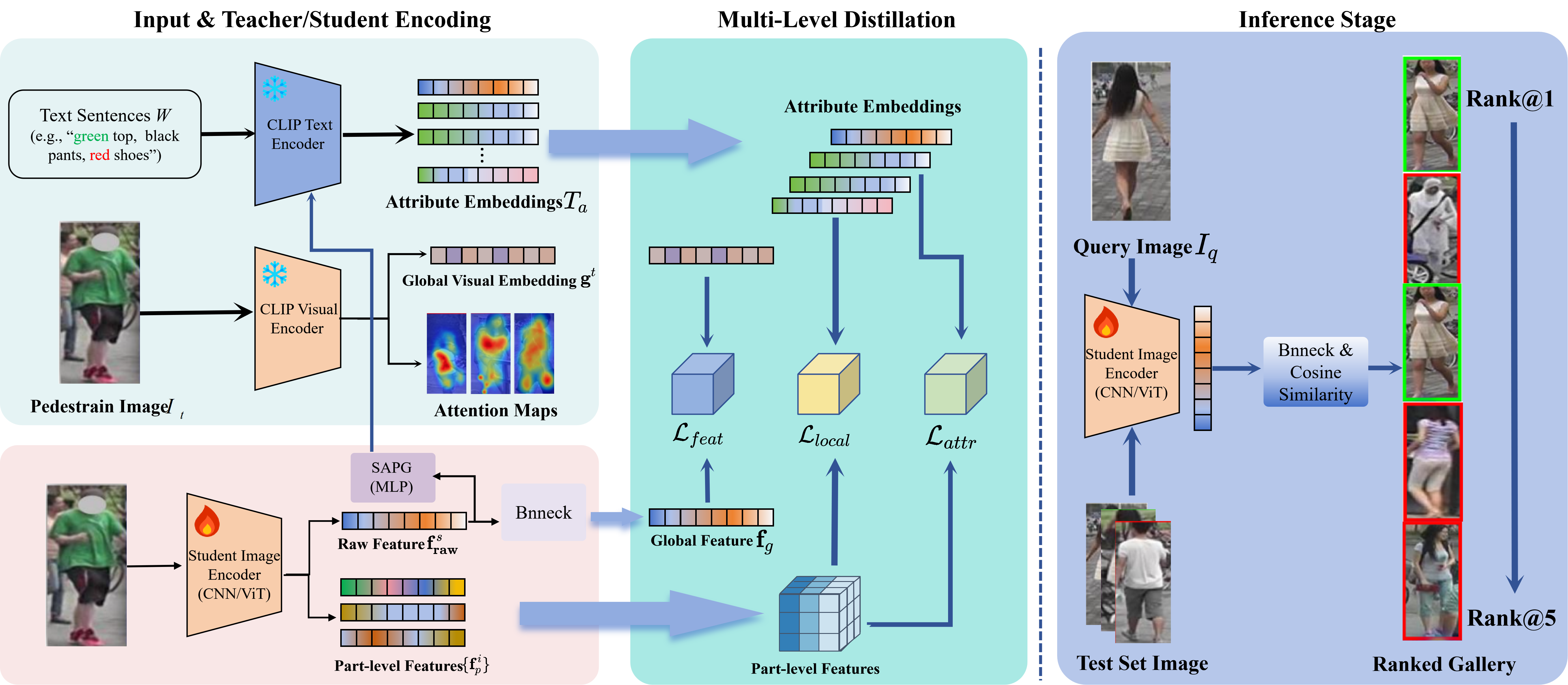}
     \caption{\fontsize{8pt}{9pt}\selectfont
        Overall pipeline of the proposed multi-level attribute--language distillation framework. 
        A frozen CLIP encoder provides global embedding $\mathbf{g}^t$, attention maps, attribute tokens $\mathbf{T}^t_a$ and local attribute tokens 
        $\mathbf{T}^t_a = \{T^t_{[x_1]},\dots,T^t_{[x_i]}\}$. 
        The student network extracts raw features $\mathbf{f}^s_{\text{raw}}$, which are supervised through 
        three branches: (1) local alignment loss $\mathcal{L}_{\text{local}}$, 
        (2) attribute semantic loss $\mathcal{L}_{\text{attr}}$, and 
        (3) global feature distillation loss $\mathcal{L}_{\text{feat}}$ guided by $\mathbf{g}^t$. 
        During inference, the Batch Normalization Neck feature $\mathbf{g}^s$ is used for cosine-similarity retrieval, 
        yielding refined ranking performance.}
    \label{fig:methodology}
\end{figure}

\section{Methodology}
\label{sec:method}

As illustrated in Fig.~\ref{fig:methodology}, our framework aims to enrich a conventional ReID backbone with fine-grained semantic cues distilled from a frozen CLIP teacher. The system operates in three stages:
(i) extracting attribute-aware textual and visual representations from CLIP, 
(ii) transferring multi-level semantics into a lightweight visual student model through global, attribute-level, and local alignment, and
(iii) performing efficient inference using only the student network without any extra expenses.

\subsection{Motivation}
Pure identity supervision in ReID methods offers only coarse signals for training. It does not guide the model to focus on useful visual details. As a result, models often ignore key attributes such as color, clothing type, or carried items. These ignored cues are important under occlusion or partial views. Without attribute guidance, models are more likely to confuse visually similar individuals.

To handle this, many ReID methods use global semantic features. These methods add attribute labels or global text cues. However, this cues just provide weak guidance for local regions, they also miss fine attribute relations across parts. As a result, their semantic signals remain coarse.

Vision–language models provide rich attribute representations through image text alignment. Their token level features highlight colors, materials, and objects. Their attention maps also indicate where these attributes appear in the image. These two properties are missing in standard ReID models. However, CLIP relies on fixed prompts and general domain data. It does not adapt well to pedestrian scenes with varied lighting or viewpoints. It is also too heavy for common ReID deployment.

Our goal is to transfer CLIP attribute cues into a compact ReID backbone. This transfer should supply attribute semantics and spatial grounding during training. It must not add extra modules during inference. Therefore, the student model  remains efficient while gaining stronger semantic awareness.
\subsection{CLIP Attribute Encoding}

We begin by extracting structured attribute semantics from CLIP~\cite{Radford2021CLIP}. A multimodal LLM produces descriptive sentences $\mathbf{W}$, each containing multiple attribute phrases that decompose pedestrian appearance into interpretable components. 

\paragraph{Textual Encoding}
The CLIP text encoder transforms $\mathbf{W}$ into a sequence of token embeddings:
\begin{equation}
\mathbf{T}^t = \{\mathbf{T}^t_{[x_1]}, \dots, \mathbf{T}^t_{[x_L]}\}.
\end{equation}
Each token embedding corresponds to a specific word or subword unit. Tokens for colors, clothing types, or accessories are grouped into attribute embeddings $\mathbf{T}^t_a$. Each embedding represents one attribute and provides a clear target for the student during distillation. These embeddings define a stable, high-level representation space that is less sensitive to image noise, lighting conditions, or local artifacts.

\paragraph{Visual Encoding}
The CLIP visual encoder provides two forms of supervision: a global visual embedding $\mathbf{g}^t$ capturing holistic semantics, and token-to-patch attention maps that indicate how each textual token attends to specific visual regions.
These attention maps provide a soft spatial prior for where particular attributes are likely to appear (Fig.~\ref{fig:methodology}, left), enabling the student model to learn not only what attributes exist but also where they are grounded in the image.
This dual encoding semantic separation through textual tokens and spatial grounding through visual attention addresses two long-standing challenges in ReID: recognizing subtle appearance differences and resolving spatial ambiguity under occlusion.

Although MLLM-generated descriptions offer rich attribute semantics without manual annotation, they may contain errors under adverse conditions such as occlusion or low illumination. To mitigate the impact of noisy descriptions, our framework leverages the frozen CLIP teacher’s attention maps as a spatial filter. Only textual attributes with strong visual grounding are used for distillation. Attributes with weak or diffuse attention are down-weighted during local alignment, effectively acting as an implicit reliability filter. This design ensures that the student receives supervision primarily from visually supported attributes, enhancing robustness to MLLM hallucination or inaccuracies.
\subsection{Multi-Level Attribute--Language Distillation}

We employ a lightweight ReID model (ResNet-IBN or ViT), which outputs a global feature $\mathbf{f}_\mathbf{g}$ and part-level representations $\{\mathbf{f}_\mathbf{p}^i\}_{i=1}^P$. Here, $P$ denotes the total number of body parts (local regions) used for part-level feature extraction. Semantic transfer occurs at three complementary levels: global, attribute, and local. These features ensure comprehensive alignment with the teacher (Fig.~\ref{fig:methodology}, middle).

\subsubsection{Scene-Aware Prompt Generator (SAPG)}

Although LLM-generated descriptions capture rich semantics, they may not always align with the actual visual content, especially under occlusion or adverse lighting. To mitigate such mismatch, we introduce a soft prompt computed from the student’s global feature:

\begin{equation}
\mathbf{p} = \mathrm{MLP}(\mathbf{f}_g)
\end{equation}
and insert it as prefix tokens:
\begin{equation}
[\mathbf{p}, \mathbf{W}] \rightarrow \mathrm{CLIP~Text~Encoder}.
\end{equation}

The soft prompt functions as an adaptive correction mechanism. It guides CLIP to emphasize visually relevant tokens, suppress noisy or unsupported descriptions, and adjust textual embeddings toward the distribution of surveillance images. SAPG is jointly optimized with the student, ensuring consistent cross-modal interpretation throughout training.

\subsubsection{Global Semantic Distillation}

To stabilize the embedding space and reduce modality drift, we align the global features of the student and teacher:

\begin{equation}
\mathcal{L}_{\text{feat}} = 1 - \cos(\mathbf{f}_{\text{g}}, \mathbf{g}^t)\end{equation}

This encourages the student to capture high-level semantics beyond identity labels and improves its baseline discriminative capability.

\subsubsection{Attribute-Level Distillation}

Fine-grained recognition depends heavily on disentangling subtle attribute cues. We therefore align student attribute features $\mathbf{f}^s_{a,i}$ with teacher attribute embeddings $\mathbf{T}^t_{a,i}$ using a contrastive objective:

\begin{equation}
    \mathcal{L}_{\text{attr}} = -\!\sum_{a,i}\log \frac{\exp(\mathbf{f}^s_{a,i}\cdot\mathbf{T}^t_{a,i}/\tau) }{ \sum_j \exp(\mathbf{f}^s_{a,i}\cdot\mathbf{T}^t_{a,j}/\tau) }
\end{equation}

This alignment performed for each attribute category. It encourages the student to internalize semantic distinctions consistently across varied environments, improving robustness to illumination, viewpoint, and camera differences.

\subsubsection{Local Attribute--Region Alignment}

Attribute semantics must be correctly localized, not merely recognized globally. For each attribute token, we use CLIP attention to identify the top-$K$ most relevant patches. These patches are averaged to form a region embedding $\mathbf{l}^{t}_{a,p}$, which supervises the student’s part representations:
\begin{equation}
\mathcal{L}_{local} = 1 - \cos(\mathbf{f}_p^s, \mathbf{l}_{a,p}^t)
\end{equation}
This spatial alignment enables the student to associate attributes with typical human-body regions, thereby reducing ambiguity under partial visibility and improving matching consistency across cameras.

\subsubsection{Progressive Optimization Strategy}

We use a coarse-to-fine schedule to reduce conflict between objectives. Global alignment is applied first to build a stable feature space.
Attribute-level supervision follows once global features are reliable.
Local alignment is added last to guide spatial grounding on top of learned attributes.

\subsubsection{Discriminative ReID Losses}
Before applying semantic distillation, the student is trained with standard ReID objectives. The identity loss supervises class-level discrimination,
\begin{equation}
\mathcal{L}_{\text{ID}} = - \sum_{i} \log p(y_i \mid \mathbf{f}_g),
\end{equation}
while the triplet loss enforces margin-based separation between samples,
\begin{equation}
\mathcal{L}_{\text{Triplet}} = 
\max\big(0,\; d(\mathbf{f}_g^a,\mathbf{f}_g^p) - d(\mathbf{f}_g^a,\mathbf{f}_g^n) + m \big).
\end{equation}
Here, $y_i$ denotes the identity label of sample $i$. In the triplet loss, $a$, $p$ and $n$ refer to the anchor, positive, and negative samples.
Together, these losses build a strong discriminative foundation before integrating attribute-level and local supervision.

\subsubsection{Overall objective}

The full loss combines discriminative and semantic objectives:
\begin{equation}
\mathcal{L}_{\text{total}}
= \mathcal{L}_{\text{ID}} + \mathcal{L}_{\text{Triplet}}
+\lambda_{\text{feat}}\mathcal{L}_{\text{feat}}
+\lambda_{\text{attr}}\mathcal{L}_{\text{attr}}
+\lambda_{\text{local}}\mathcal{L}_{\text{local}}
\end{equation}
The identity and triplet losses ensure the model preserves strong visual discrimination. The global, attribute, and local distillation losses supply complementary semantic signals that enhance feature quality at different levels. Together, these terms guide the student to learn stable global structure, meaningful attribute semantics, and consistent spatial grounding, forming a unified representation that remains efficient at inference.
During inference, the MLLM and CLIP teacher are discarded. The SAPG module utilizes a lightweight MLP to generate prompts directly from the student's visual features $\mathbf{f}_g$, ensuring zero extra computational overhead from heavy models at test time.

\section{Experiments}
\subsection{Experiment Settings}
\subsubsection{Datasets and evaluation metrics}
For a fair comparison with previous person Re-Identification (ReID) works, we adopt three standard benchmarks: Market-1501, DukeMTMC-ReID, and MSMT17.
Evaluation follows the conventional single-query setting with Rank@K (R@K) and mean Average Precision (mAP) as metrics.
Higher values of R@K and mAP indicate better retrieval performance.
Market-1501\cite{zheng2015scalable} contains 1,501 identities, 12,936 training images, and 19,732 testing images captured from 6 cameras. DukeMTMC-ReID\cite{ristani2016MTMC} provides 1,404 identities with 16,522 training images and 19,889 testing images under 8 cameras. MSMT17\cite{Wei_2018_CVPR} is a large-scale and challenging dataset collected under 15 cameras across both indoor and outdoor scenes.
It contains 4,101 identities and 126,441 images captured at different times of day (morning, noon, afternoon), providing rich illumination and scene diversity.

To provide fine-grained semantic supervision, we employ a MLLM (Qwen-VL) to automatically generate structured attribute prompts. Each prompt is parsed into five semantic groups: top, bottom, shoes, carrying item, and accessory to guide cross-modal distillation. All datasets are preprocessed with bounding boxes and pseudo textual descriptions generated by Qwen-VL, providing consistent attribute annotations.

\subsubsection{Implementation Details}
We adopt CLIP (ViT-B/16) as a frozen teacher and train lightweight students (ResNet-50, ResNet-IBN~\cite{pan2018IBN-Net}, or CLIP-ViT) with multi-level distillation. 
Images are resized to $224\times224$ and sampled via the P--K strategy (P=16, K=4). 
Training uses AdamW with OneCycle scheduling for 120 epochs; learning rates are $3.5\times10^{-4}$ for non-CLIP parameters and $1\times10^{-5}$ for CLIP modules. We enable  
progressive distillation at epochs 20/40/80, which gradually introduces global, attribute, and local alignment to stabilize optimization.  
The student relies on BN-neck features during testing.  
Horizontal flip augmentation is applied, and k-reciprocal Re-ranking is optionally used for the final retrieval accuracy.In Equation (9), the balancing weights are empirically set as $\lambda_{\text{feat}} = 1.0, \lambda_\text{attr}= 0.5, \lambda_\text{local} = 0.1$.

\begin{table}[t]
\centering
\scriptsize
\setlength{\tabcolsep}{2.6pt}  
\renewcommand{\arraystretch}{1.05}
\caption{Comparison with state-of-the-art person ReID methods on three benchmarks. Best results are in \textbf{bold}.All the results are without Re-ranking.}
\begin{tabular}{l l l cc cc cc}
\toprule
\multirow{2}{*}{Backbone} & \multirow{2}{*}{Method} & \multirow{2}{*}{Ref.} 
& \multicolumn{2}{c}{Market1501} & \multicolumn{2}{c}{MSMT17} & \multicolumn{2}{c}{DukeMTMC} \\
\cmidrule(lr){4-5}\cmidrule(lr){6-7}\cmidrule(lr){8-9}
 &  &  & mAP & R-1 & mAP & R-1 & mAP & R-1 \\
\midrule
\multirow{10}{*}{\textbf{ResNet50}} 
 & IBN-Net\cite{pan2018IBN-Net} & ECCV’18 & 88.2 & 95.0 & - & - &79.1 & 90.1\\ 
 & OSNet\cite{zhou2019osnet} & ICCV’19 & 84.9 & 94.8 & 52.9 & 78.7 & 73.5 & 88.6 \\
 & CDNet\cite{DBLP:conf/cvpr/LiWZ21} & CVPR’21 & 86.0 & 95.1 & 54.7 & 78.9 & 76.8 & 88.6 \\
 & CAL\cite{rao2021counterfactual} & ICCV’21 & 87.0 & 94.5 & 56.2 & 79.5 & 76.4 & 87.6 \\
 & ALDER\cite{DBLP:journals/tip/ZhangLFX22} & TIP’21 & 88.9 & 95.6 & 59.1 & 82.5 & 78.9 & 89.9 \\
 & AMD\cite{chen2021AMD} & ICCV’21 & 87.2 & 94.8 & - & - & 71.7 & 86.7 \\
 & CLIP3DReID\cite{Liu2024CLIP3DReID} & CVPR’24 & 88.4 & 95.6 & 61.2 & 81.5 & - & - \\
 & Baseline & - & 88.1 & 94.7 & 60.7 & 82.1 & 79.3 & 88.6 \\
 & \textbf{ALADIN (Ours)} & - & \textbf{89.4} & \textbf{96.0} & \textbf{62.0} & \textbf{83.3} & \textbf{80.4} & \textbf{91.0} \\
 \midrule
\multirow{8}{*}{\textbf{ViT-B/16}} 
 & \multicolumn{8}{l}{\textit{ViT-based method}} \\
 & TransReID\cite{He_2021_ICCV} & ICCV’21 & 88.9 & 95.2 & 67.4 & 85.3 & 82.0 & 90.7 \\
 & DCAL\cite{DBLP:conf/cvpr/ZhuKLLTS22} & CVPR’22 & 87.5 & 94.7 & 64.0 & 83.1 & 80.1 & 89.0 \\
 & PHA\cite{Zhang_2023_CVPR} & CVPR’23 & 90.2 & 96.1 & 68.9 & 86.1 & - & - \\
 & \multicolumn{8}{l}{\textit{CLIP-based method}} \\
 & CLIP-ReID\cite{DBLP:conf/aaai/LiSL23} & AAAI’23 & 89.6 & 95.5 & 73.4 & 88.7 & 82.5 & 90.0 \\
 & Baseline & - & 89.0 & 95.2 & 67.2 & 85.4 & 82.0 & 90.6 \\
 & \textbf{ALADIN (Ours)} & - & \textbf{91.1} & \textbf{96.1} & 68.8 & 86.5 & \textbf{83.3} & \textbf{91.7} \\
\bottomrule
\end{tabular}
\label{tab:sota}
\end{table}

\subsection{Comparison with State-of-the-art Methods}
Table~\ref{tab:sota} summarizes results on all three datasets.  
Under the ResNet-IBN setting, ALADIN consistently outperforms recent CNN-based ReID approaches such as CAL, ALDER, and CDNet.  
The gains on MSMT17 (62.0\% mAP, 83.3\% Rank-1) are particularly notable, indicating the advantage of attribute-level cues in challenging illumination and cross-camera scenarios.

With the ViT-B/16 backbone, ALADIN further improves upon strong transformer- and CLIP-based baselines.  
Compared with TransReID and CLIP-ReID, our method achieves higher performance on both Market1501 and DukeMTMC, demonstrating that structured attribute–language distillation enhances global semantics and patch-level discrimination simultaneously.  
These improvements validate the benefit of integrating CLIP’s attribute grounding into a deployable ReID backbone.

\subsection{Ablation Study}

\begin{table}[t]
\centering
\small
\setlength{\tabcolsep}{4pt}
\caption{Ablation on Market-1501 (single-query, without re-ranking). 
We add components progressively.}
\begin{tabular}{lcc}
\toprule
Method & mAP (\%) & Rank-1 (\%) \\
\midrule
Baseline (ID+Triplet) &  88.2 & 95.0 \\
+ Global Distill       &  88.6 & 95.3 \\
+ Mid-layer Distill    &  88.7 & 95.4  \\
+ Attribute Semantic   &  89.0 & 95.6 \\
+ Local Align          &  89.2 & 95.7 \\
+ Attribute CE         &  89.4 & 95.8 \\
\textbf{Ours (Full)}   &  \textbf{89.4} & \textbf{96.0}\\
\bottomrule
\end{tabular}
\label{tab:ablation}
\vspace{-2mm}
\end{table}

The ablation study on Market-1501 (Table~\ref{tab:ablation}) verifies the contribution of each component. Each component contributes incrementally, with full model achieving +1.2\% / +1.0\% Rank-1 over baseline.  

We also analyze the training-stability components.  
Removing SIE slightly reduces cross-camera robustness, while omitting OLP weakens intra-batch structure modeling.  
Center regularization offers small but stable improvements.

\begin{table}[t]
\centering
\caption{Impact of MLLM attribute noise on ReID performance. ``Drop'' removes (filters) attribute fields by replacing them with \texttt{unknown}; ``Wrong'' flips attribute values within the same key.}
\label{tab:mllm_noise}
\setlength{\tabcolsep}{6pt}
\begin{tabular}{l c c c c}
\toprule
\multirow{2}{*}{Setting} &
\multirow{2}{*}{Noise Type} &
\multirow{2}{*}{Ratio} &
\multicolumn{2}{c}{Market-1501} \\
\cmidrule(lr){4-5}
& & & mAP & R1 \\
\midrule
\textbf{ALADIN(Ours)} & None & 0.0 & 89.4 & 96.0 \\
\midrule
Filter-$r$ & Drop (field removal) & 0.2 & 89.2 & 95.8 \\
Filter-$r$ & Drop (field removal) & 0.4 & 88.8 & 95.2 \\
Filter-$r$ & Drop (field removal) & 0.6 & 88.4 & 95.0 \\
Filter-$r$ & Drop (field removal) & 0.8 & 88.2 & 94.9 \\
\midrule
Wrong-$p$ & Wrong (all keys) & 0.1 & 88.7 & 95.3 \\
Wrong-$p$ & Wrong (all keys) & 0.2 & 88.0 & 94.4 \\
Wrong-$p$ & Wrong (all keys) & 0.3 & 87.4 & 93.3 \\
Wrong-$p$ & Wrong (all keys) & 0.4 & 86.5 & 92.1 \\
\bottomrule
\end{tabular}
\end{table}

\subsubsection{Analysis of MLLM Prompts}
\label{tab:MLLM}
This section analyzes the robustness of ALADIN to noise in MLLM-generated attribute prompts. The proposed framework relies on textual attributes to guide attribute-level and local cross-modal alignment.
Therefore, inaccuracies in MLLM-generated prompts can degrade the quality of semantic supervision. We therefore conduct controlled perturbation experiments to systematically evaluate the impact of attribute noise.

\paragraph{Impact of Attribute Filtering and Corruption.}
We first study how different types and levels of noise in MLLM-generated attributes influence ReID performance. As shown in Table~\ref{tab:mllm_noise}, we consider two representative noise sources: \emph{attribute filtering} (Drop), where a portion of attribute fields is removed by replacing them with \texttt{unknown}, and \emph{attribute corruption} (Wrong), where attribute values are replaced with incorrect ones within the same semantic category. The noise ratio controls the proportion of affected attributes.

When no noise is introduced, ALADIN achieves a baseline performance of (mAP= 89.4\%, R1= 96.0\%) on Market-1501. As the filtering ratio $r$ increases from 0.2 to 0.8, performance gradually degrades, indicating that missing attribute information weakens semantic guidance but does not completely collapse the learned representation. Notably, moderate filtering levels (e.g., $r=0.2$ or $r=0.4$) still preserve competitive accuracy, suggesting that ALADIN is robust to partial attribute omission.

In contrast, attribute corruption introduces more severe performance drops. As the wrong-attribute ratio $p$ increases, both mAP and Rank-1 accuracy decrease more rapidly compared with the filtering case. This observation highlights that incorrect semantic supervision is more harmful than missing supervision, as erroneous attributes may mislead the cross-modal alignment process and encourage inconsistent feature learning.

\paragraph{Effect of Different Attribute Error Types.}
To further understand which types of attribute errors are most detrimental, we analyze the impact of different corruption sources under a fixed wrong ratio ($p=0.2$). Table~\ref{tab:mllm_error_type} reports results for color errors, category errors (top, bottom, and shoes), existence errors (carrying objects and accessories), and a mixed setting.
Specifically, category errors refer to incorrect clothing-type assignments for top, bottom, or shoes (e.g., mislabeling a jacket as a T-shirt), color errors correspond to incorrect color descriptions, and existence errors indicate whether carried objects or accessories are incorrectly marked as present or absent.

Among the evaluated error types, color and category errors lead to noticeable performance degradation (mAP= 0.2\% /R1=0.3\% and mAP= 0.4\% / R1= 0.5\%, respectively), reflecting the importance of fine-grained appearance attributes in person re-identification. Existence errors exhibit a relatively smaller impact, while the mixed-error setting yields the largest degradation, confirming that compound semantic inconsistencies are particularly harmful.

Overall, these results demonstrate that ALADIN remains robust under moderate prompt noise, while emphasizing the importance of maintaining attribute correctness when using MLLMs for semantic supervision.

\begin{table}[t]
\centering
\caption{Breakdown of wrong-attribute types at a fixed wrong ratio $p = 0.2$ on Market-1501.}
\label{tab:mllm_error_type}
\setlength{\tabcolsep}{6pt}
\begin{tabular}{l c c c}
\toprule
Error Type & $p$ & mAP & R1 \\
\midrule
Color error only & 0.2 & 89.0 & 95.5 \\
Category error only (top/bottom/shoes) & 0.2 & 88.8 & 95.3 \\
Existence error only (carrying/accessory) & 0.2 & 88.7 & 95.1 \\
Mixed (all above) & 0.2 & 88.0 & 94.4 \\
\bottomrule
\end{tabular}
\end{table}

\begin{table}[!t]
\centering
\caption{Ablation of soft prompt strategies on Market-1501. Fixed prompts are shared across samples; scene-aware prompts are conditioned on the student feature.}
\label{tab:soft_prompt_ablation}
\setlength{\tabcolsep}{4pt}
\begin{tabular}{l c c c}
\toprule
Variant & Prompt Type & mAP & R1 \\
\midrule
w/o Soft Prompt & None & 88.7 & 95.5 \\
Random Prompt (control) & Random, not trained & 88.5 & 95.4 \\
Fixed Soft Prompt & Learnable, global & 89.0 & 95.7 \\
Scene-Aware Soft Prompt (Ours) & Conditioned, per-sample & 89.4 & 96.0 \\
\bottomrule
\end{tabular}
\end{table}

\subsubsection{Ablation Study of SAPG}
\label{tab:SAPG}
This section evaluates the effectiveness of the proposed Scene-Aware Prompt Generator (SAPG) through a series of ablation studies. The goal is to isolate the contribution of soft prompts and to analyze how different prompt designs influence cross-modal distillation and ReID performance.

\paragraph{Effect of Soft Prompt Strategies.}
We first compare different soft prompt strategies, as summarized in Table~\ref{tab:soft_prompt_ablation}. The baseline without soft prompts directly encodes attribute templates using the frozen CLIP text encoder, yielding performance of (mAP= 88.7\%, R1=95.5\%) on Market-1501. Introducing random, non-trainable prompts results in comparable or slightly degraded performance, indicating that simply adding extra tokens does not provide meaningful semantic benefits.

In contrast, incorporating learnable fixed soft prompts leads to consistent improvements across Market-1501, demonstrating that prompt adaptation can enhance text–visual alignment. The proposed scene-aware soft prompts further improve performance, achieving the best results with (mAP=89.4\%, R1=96.0\%) on Market-1501. This gain suggests that conditioning prompts on the student visual features enables the teacher text encoder to produce more image-specific semantic representations, thereby facilitating more effective attribute-level and local distillation.

\paragraph{Effect of Soft Prompt Length.}
We further investigate the influence of the number of soft prompt tokens $N$, with results reported in Table~\ref{tab:soft_prompt_length}. Increasing $N$ from 2 to 4 yields noticeable performance gains, indicating that a small number of prompt tokens is sufficient to inject additional contextual information. However, further increasing $N$ to 8 results in marginal improvements or performance saturation, suggesting diminishing returns when the prompt becomes overly long.

Based on these observations, we adopt $N=4$ as the default setting in all experiments, which achieves a favorable trade-off between performance and model complexity.

\begin{table}[t]
\centering
\caption{Effect of soft prompt length $N$ (number of soft tokens) on Market-1501.}
\label{tab:soft_prompt_length}
\setlength{\tabcolsep}{6pt}
\begin{tabular}{c c c}
\toprule
Prompt Length $N$ & mAP & R1 \\
\midrule
2 & 89.1 & 95.8 \\
4 & 89.4 & 96.0 \\
8 & 89.5 & 96.2 \\
\bottomrule
\end{tabular}
\end{table}

\subsection{Cross-domain Evaluation}
To further investigate the generalization ability of the proposed method under domain shift, we conduct cross-domain experiments between Market-1501 (M) and DukeMTMC-ReID (D). Specifically, the model is trained on one dataset and directly evaluated on the other without any fine-tuning. As shown in Table~\ref{tab:cross_domain}, compared with the baseline ResNet-50 and the existing attribute-based explanation method AMD~\cite{chen2021AMD}, our method consistently improves both Rank-1 accuracy and mAP in all transfer settings.

For the M$\rightarrow$D scenario, our approach achieves 3.18\% and 3.99\% absolute improvements in Rank-1 and mAP over the baseline, respectively, demonstrating that the attribute-aware semantic alignment in our framework effectively reduces the cross-domain discrepancy. Similarly, under the D$\rightarrow$M setting, our method yields +2.75\% Rank-1 and +2.07\% mAP gains, further validating the robustness of the proposed model when transferring to unseen target domains.

These results confirm that leveraging cross-modal attribute semantics not only enhances interpretability but also improves domain generalization, indicating the potential of our strategy to bridge distribution gaps across different surveillance environments.

\begin{table}[t]
\centering
\caption{Cross-domain evaluation results of ResNet-50 and AMD. 
M\textrightarrow D: trained on Market-1501 and tested on DukeMTMC-ReID; 
D\textrightarrow M: reverse setting.}
\label{tab:cross_domain}

\begin{tabular}{l l c c c}
\toprule
Datasets & Models & Rank-1 (\%) & Rank-5 (\%) & mAP (\%) \\
\midrule
\multirow{2}{*}{M $\rightarrow$ D} 
& ResNet-50 & 45.02 & 61.40 & 26.43 \\
& AMD\cite{chen2021AMD} & 47.08 & 62.61 & 28.41 \\
& \textbf{Ours} & 48.20 & 63.95 & 29.42 \\
\midrule
\multirow{2}{*}{D $\rightarrow$ M} 
& ResNet-50 & 53.15 & 71.26 & 24.48 \\
& AMD\cite{chen2021AMD} & 54.48 & 71.59 & 25.48 \\
& \textbf{Ours} & 55.90 & 72.30 & 26.55 \\
\bottomrule
\end{tabular}
\end{table}

\begin{figure*}[t]
    \centering
    \includegraphics[width=0.7
    \linewidth]{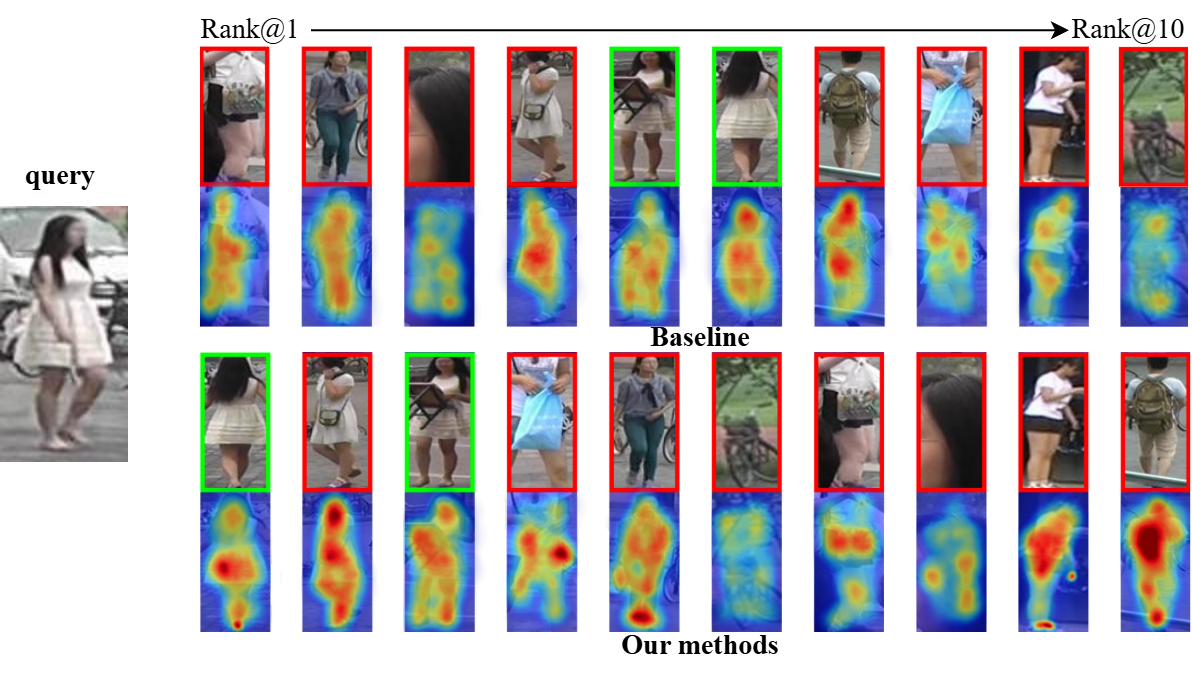}
    \caption{\fontsize{8pt}{9pt}\selectfont
    Qualitative comparison of the same retrieval batch under baseline and our method. 
    Both rows display identical gallery images but with distinct attention patterns. 
    The baseline exhibits uniform, global attention across the entire person region, 
    lacking fine-grained discrimination. 
    In contrast, our method is designed to focus on semantically meaningful local attributes,such as clothing texture, color, and accessories,
    enabling more discriminative matching and improved interpretability. 
    This shift from global to attribute-aware attention contributes to higher retrieval accuracy in challenging cases.}
    \label{fig:qualiative}
\end{figure*}

\subsection{Qualitative Results}
Fig.~\ref{fig:qualiative} visualizes retrieval examples.  
The baseline mainly focuses on coarse global appearance, often ranking visually similar but semantically mismatched candidates.  
ALADIN , instead, emphasizes CLIP-guided attribute cues such as color, clothing type, and accessories, allowing it to retrieve identity-consistent results with improved interpretability.  
This confirms that attribute–language distillation leads to more reliable instance-level matching under occlusion, viewpoint variation, and distractor scenarios.

\section{Conclusion}

This work demonstrates that fine-grained attribute--language distillation from a frozen CLIP model can effectively enhance lightweight person ReID.
Through attribute-level contrastive supervision, attention-guided local alignment, and scene-aware prompt refinement, the student learns richer and more robust representations without added inference cost.
Experiments on Market-1501, DukeMTMC-ReID, and MSMT17 demonstrate consistent gains in accuracy and interpretability, with ablations confirming the contribution of each component.

%
%
%

\begin{thebibliography}{8}
\bibitem{Luo_2019_CVPR_Workshops}
H.~Luo, Y.~Gu, X.~Liao, S.~Lai, and W.~Jiang, ``Bag of tricks and a strong baseline for deep person re-identification,'' in \emph{IEEE Conference on Computer Vision and Pattern Recognition Workshops (CVPRW)}, 2019.

\bibitem{sun2018PCB}
Y.~Sun, L.~Zheng, Y.~Yang, Q.~Tian, and S.~Wang, ``Beyond part models: Person retrieval with refined part pooling (and a strong convolutional baseline),'' in \emph{ECCV}, 2018.

\bibitem{He_2021_ICCV}
S.~He, H.~Luo, P.~Wang, F.~Wang, H.~Li, and W.~Jiang, ``Transreid: Transformer-based object re-identification,'' in \emph{Proceedings of the IEEE/CVF International Conference on Computer Vision (ICCV)}, 2021, pp. 15\,013--15\,022.

\bibitem{Radford2021CLIP}
A.~Radford, J.~W. Kim, C.~Hallacy, A.~Ramesh, G.~Goh, S.~Agarwal, G.~Sastry, A.~Askell, P.~Mishkin, J.~Clark, G.~Krueger, and I.~Sutskever, ``Learning transferable visual models from natural language supervision,'' in \emph{Proceedings of the 38th International Conference on Machine Learning (ICML)}, 2021, pp. 8748--8763.

\bibitem{DBLP:conf/aaai/LiSL23}
S.~Li, L.~Sun, and Q.~Li, ``{CLIP-ReID: Exploiting Vision-Language Model for Image Re-identification without Concrete Text Labels},'' in \emph{Proceedings of the AAAI Conference on Artificial Intelligence (AAAI)}, 2023, pp. 1405--1413.

\bibitem{Yang2024PromptSG}
Z.~Yang, Y.~Zhang, Y.~Rao, J.~Zhou, and J.~Lu, ``A pedestrian is worth one prompt: Towards language-guided person re-identification,'' in \emph{Proceedings of the IEEE/CVF Conference on Computer Vision and Pattern Recognition (CVPR)}, 2024, pp. 17\,291--17\,300.

\bibitem{Liu2024CLIP3DReID}
F.~Liu, Z.~Wei, S.~Lu, and X.~Li, ``Distilling clip with dual guidance for learning discriminative human body shape representation,'' in \emph{Proceedings of the IEEE/CVF Conference on Computer Vision and Pattern Recognition (CVPR)}, 2023, pp. 15\,130--15\,139.

\bibitem{wang2025idea}
Y.~Wang, Y.~Lv, P.~Zhang, and H.~Lu, ``Idea: Inverted text with cooperative deformable aggregation for multi-modal object re-identification,'' in \emph{Proceedings of the IEEE/CVF Conference on Computer Vision and Pattern Recognition (CVPR)}, 2025.

\bibitem{Qwen-VL}
J.~Bai, S.~Bai, S.~Yang, S.~Wang, S.~Tan, P.~Wang, J.~Lin, C.~Zhou, and J.~Zhou, ``Qwen-vl: A frontier large vision-language model with versatile abilities,'' \emph{arXiv preprint arXiv:2308.12966}, 2023.

\bibitem{zhang2024clipvisinfra}
D.~Zhang, H.~Chen, Y.~Xu, W.~Li, and Y.~Wu, ``Clip-driven semantic discovery network for visible–infrared re-identification,'' \emph{arXiv preprint arXiv:2401.05806}, 2024.

\bibitem{zhou2022coop}
K.~Zhou, J.~Yang, C.~C. Loy, and Z.~Liu, ``Learning to prompt for vision-language models,'' \emph{International Journal of Computer Vision (IJCV)}, 2022.

\bibitem{DBLP:conf/eccv/SomersAV24}
V.~Somers, A.~Alahi, and C.~D. Vleeschouwer, ``Keypoint promptable re-identification,'' in \emph{European Conference on Computer Vision (ECCV)}, 2024, pp. 216--233.

\bibitem{li2023prototypical}
J.~Li and X.~Gong, ``Prototypical contrastive learning-based clip fine-tuning for object re-identification,'' \emph{arXiv preprint arXiv:2310.17218}, 2023.

\bibitem{chen2021AMD}
L.~Chen, Z.~Liu, and M.~Zhang, ``Explainable person re-identification with attribute-guided metric distillation,'' in \emph{IEEE International Conference on Computer Vision (ICCV)}, 2021.

\bibitem{hao2021cmreid}
Y.~Hao, G.-J. Zhang, Y.~Wang, and Z.~Sun, ``Cross-modality person re-identification via modality confusion and center aggregation,'' in \emph{Proceedings of the IEEE/CVF International Conference on Computer Vision (ICCV)}, 2021, pp. 19\,822--19\,831.

\bibitem{DBLP:conf/cvpr/ZhuKLLTS22}
H.~Zhu, W.~Ke, D.~Li, J.~Liu, L.~Tian, and Y.~Shan, ``Dual cross-attention learning for fine-grained visual categorization and object re-identification,'' in \emph{{IEEE/CVF} Conference on Computer Vision and Pattern Recognition, {CVPR} 2022, New Orleans, LA, USA, June 18-24, 2022}.\hskip 1em plus 0.5em minus 0.4em\relax {IEEE}, 2022, pp. 4682--4692.

\bibitem{yan2023tipclip}
C.~Yan, Y.~Zhang, J.~Chen, and C.~Shen, ``Clip-driven fine-grained text–image person re-identification (tip-clip),'' \emph{IEEE Transactions on Image Processing}, pp. 2525--2537, 2023.

\bibitem{shao2023unipt}
Z.~Shao, J.~Zhao, B.~Li, C.~Song, H.~Tang, Y.~Yan, and L.~Shao, ``Unified pre-training with pseudo texts for text-to-image person re-identification (unipt),'' in \emph{Proceedings of the IEEE/CVF International Conference on Computer Vision (ICCV)}, 2023, pp. 10\,103--10\,112.

\bibitem{tan2024harnessing}
W.~Tan, C.~Ding, J.~Jiang, F.~Wang, Y.~Zhan, and D.~Tao, ``Harnessing the power of mllms for transferable text-to-image person reid,'' \emph{CVPR}, 2024.

\bibitem{Zhang_2024_CVPR}
P.~Zhang, Y.~Wang, Y.~Liu, Z.~Tu, and H.~Lu, ``Magic tokens: Select diverse tokens for multi-modal object re-identification,'' in \emph{Proceedings of the IEEE/CVF Conference on Computer Vision and Pattern Recognition (CVPR)}, 2024.

\bibitem{yu2024tfclip}
C.~Yu, Z.~Chen, H.~Zhang, J.~Wang, and M.~Tang, ``Tf-clip: Learning text-free clip for video-based person re-identification,'' in \emph{Proceedings of the AAAI Conference on Artificial Intelligence (AAAI)}, 2024, pp. 16\,742--16\,750.

\bibitem{pan2018IBN-Net}
P.~Xingang, L.~Ping, S.~Jianping, and T.~Xiaoou, ``Two at once: Enhancing learning and generalization capacities via ibn-net,'' in \emph{ECCV}, 2018.

\bibitem{zhou2019osnet}
K.~Zhou, Y.~Yang, A.~Cavallaro, and T.~Xiang, ``Omni-scale feature learning for person re-identification,'' in \emph{ICCV}, 2019.

\bibitem{DBLP:conf/cvpr/LiWZ21}
H.~Li, G.~Wu, and W.~Zheng, ``Combined depth space based architecture search for person re-identification,'' in \emph{{IEEE} Conference on Computer Vision and Pattern Recognition, {CVPR} 2021, virtual, June 19-25, 2021}.\hskip 1em plus 0.5em minus 0.4em\relax Computer Vision Foundation / {IEEE}, 2021.

\bibitem{rao2021counterfactual}
Y.~Rao, G.~Chen, J.~Lu, and J.~Zhou, ``Counterfactual attention learning for fine-grained visual categorization and re-identification,'' in \emph{ICCV}, 2021.

\bibitem{DBLP:journals/tip/ZhangLFX22}
Q.~Zhang, J.~Lai, Z.~Feng, and X.~Xie, ``Seeing like a human: Asynchronous learning with dynamic progressive refinement for person re-identification,'' \emph{{IEEE} Trans. Image Process.}, pp. 352--365, 2022.

\bibitem{Zhang_2023_CVPR}
Z.~Guiwei, Z.~Yongfei, Z.~Tianyu, L.~Bo, and P.~Shiliang, ``Pha: Patch-wise high-frequency augmentation for transformer-based person re-identification,'' in \emph{Proceedings of the IEEE Conference on Computer Vision and Pattern Recognition (CVPR)}, 2023, pp. 14\,133--14\,142.

\bibitem{zheng2015scalable}
L.~Zheng, L.~Shen, L.~Tian, S.~Wang, J.~Wang, and Q.~Tian, ``Scalable person re-identification: A benchmark,'' in \emph{Computer Vision, IEEE International Conference on}, 2015.

\bibitem{ristani2016MTMC}
E.~Ristani, F.~Solera, R.~Zou, R.~Cucchiara, and C.~Tomasi, ``Performance measures and a data set for multi-target, multi-camera tracking,'' in \emph{European Conference on Computer Vision workshop on Benchmarking Multi-Target Tracking}, 2016.

\bibitem{Wei_2018_CVPR}
L.~Wei, S.~Zhang, W.~Gao, and Q.~Tian, ``Person transfer gan to bridge domain gap for person re-identification,'' in \emph{Proceedings of the IEEE Conference on Computer Vision and Pattern Recognition (CVPR)}, 2018, pp. 79--88.
\end{thebibliography}
%

\end{document}